# Um Método para Busca Automática de Redes Neurais Artificiais

**Anderson P. da Silva[1], Teresa B. Ludermir[1], Leandro M. Almeida[1]**

[1]Centro de Informática – Universidade Federal de Pernambuco (UFPE)
Caixa Postal 50740-560 – Recife– PE – Brasil

`{aps3,tbl,lma3}@cin.ufpe.br`

***Abstract.*** *This paper describes a method that automatically searches Artificial Neural Networks using Cellular Genetic Algorithms. The main difference of this method for a common genetic algorithm is the use of a cellular automaton capable of providing the location for individuals, reducing the possibility of local minima in search space. This method employs an evolutionary search for simultaneous choices of initial weights, transfer functions, architectures and learning rules. Experimental results have shown that the developed method can find compact, efficient networks with a satisfactory generalization power and with shorter training times when compared to other methods found in the literature.*

***Resumo.*** *Este artigo descreve um método de busca automática por Redes Neurais Artificiais (RNAs) utilizando Algoritmos Genéticos Celulares (AGCs). A principal diferença entre AGCs e algoritmo genéticos (AGs) é o uso de um autômato celular para dar localização aos indivíduos, reduzindo as chances de estes caírem em mínimos locais no espaço de busca. O AGCRN-CR executa buscas evolucionárias simultâneas de pesos, funções de transferência, arquiteturas e regras de aprendizado. Resultados experimentais mostram que o AGCRN-CR pode encontrar redes compactas e eficientes com satisfatório poder de generalização e curto tempo de treinamento em relação a outros métodos encontrados na literatura.*

## 1. Introdução

A busca por redes neurais artificiais (RNAs) que se adéquem a determinados problemas é vista como uma questão importante na aplicação de RNAs, uma vez que o poder de generalização de uma rede neural pode ser melhorado de acordo com a devida escolha dos seus parâmetros. Tal busca possui uma série de dificuldades, visto que a construção de uma RNA com configuração que se aproxime do ótimo envolve questões como o exponencial número de parâmetros que precisam ser ajustados, a necessidade do conhecimento *a priori* do domínio do problema e do funcionamento da RNA para definição destes problemas, além da necessidade de um especialista quando este conhecimento não existe [Almeida and Ludermir, 2007]. Essa busca por uma configuração de RNA adequada, de forma geral, é feita manualmente ou através de métodos de busca que consigam encontrar configurações ótimas ou se aproximem do ótimo. Nesse contexto, alguns pesquisadores propõem métodos de busca evolucionária que visam "evoluir" as soluções do problema em questão a partir de uma solução inicial (ou conjunto de soluções iniciais) [Abraham, 2004, Garcia-Pedrajas et al., 2005, Carvalho and Ludermir, 2006, Gomes and Ludermir, 2013, Figueiredo and Ludermir, 2014].

Neste trabalho, apresentamos um método de busca que utiliza RNAs e um tipo diferente de AGs, conhecido como Algoritmo Genético Celular (AGC) com codificação direta. Este método é denominado AGCRN-CR (Algoritmo Genético Celular + Rede Neural + Codificação Real), e baseia-se no método proposto por [Almeida and Ludermir, 2007]. O AGCRN-CR faz busca por pesos, arquiteturas, funções de ativação e regras de aprendizagem para rede neural através de um AG comum e visa encontrar redes compactas com bom poder de generalização. A principal diferença do AGCRN-CR em relação ao NNGA-DCOD é o tipo de algoritmo genético utilizado, onde um Autômato Celular (AC) trabalha em conjunto com um Algoritmo Genético para dar localidade aos elementos do AG, formando o AGC. O restante desse artigo é organizado da seguinte forma. A Seção 2 apresenta os conceitos básicos de algoritmos genéticos enquanto a Seção 3 apresenta os conceitos de algoritmos evolucionários celulares. Em seguida a Seção 4 descreve o método AGCRN-CR e a Seção 5 descreve os resultados experimentais. Por fim, na Seção 6 encontram-se as conclusões e os trabalhos futuros.

## 2. Algoritmos Genéticos

Os algoritmos genéticos (AGs) são técnicas que visam achar soluções ótimas ou quase-ótimas em problemas de otimização e busca. AGs são inspirados em processos naturais de evolução [Linden, 2006], principalmente através de conceitos de biologia evolutiva como herança, mutação, seleção natural das espécies, e outros. Através da união de técnicas evolucionárias (entre elas AGs) e RNAs na busca dos próprios componentes das RNAs, temos as redes neurais artificiais evolucionárias (RNAEs), que foram definidas por [Yao, 1999]. As RNAEs conseguem ter melhores *exploitations* e *explorations* de grande número de aspectos e componentes necessários (como pesos, arquiteturas e funções de transição) para a construção de RNAs com bom poder de generalização.

No entanto, um dos problemas mais comuns na utilização de métodos evolucionários é a convergência da busca dentro de um ponto de mínimo local no espaço de pesquisa. O tratamento deste caso deve ser feito com maior cuidado quando se trata de uma busca evolutiva através de algoritmos genéticos, já que a evolução dos indivíduos sem o correto controle pode gerar a produção de super indivíduos que possuem sempre a mesma carga genética (ou uma carga genética muito parecida) e que irão predominar em relação a todos os outros, trazendo a convergência genética à população e reduzindo a diversidade da mesma. Se a população de um AG tiver pouca diversidade, as chances de encontrar soluções diferentes, ou até mesmo a solução ideal, é reduzida, pois a diversidade genética é essencial para que o AG encontre as soluções ideais [Almeida and Ludermir, 2007].

Ainda assim, diversos pesquisadores têm empregado os algoritmos genéticos como ferramenta de busca por espaços de ótimos globais com sucesso em suas pesquisas [Yao, 1999, Minku and Ludermir, 2008, Ferreira and Ludermir, 2009]. Nestas pesquisas foi possível perceber que apesar de possuir, em geral, bons resultados, as principais desvantagens dos AGs são as perturbações de boas sub-soluções pelas operações de cruzamento e mutação, além da perda da diversidade genética causada pelo operador de seleção, que constantemente diminui a variedade dos cromossomos [Rajabalipour, et al. 2009].

## 3. Algoritmos Evolucionários Celulares

Uma das formas encontradas para amenizar as perturbações nas soluções causadas pelo operador de seleção dos AGs foi proposta por [Cao, et al. 1998], onde os pesquisadores introduziram autômatos celulares (AC) para realizar a localização e vizinhança na estrutura da população dos algoritmos genéticos. Um AC é um modelo discreto estudado na teoria da computabilidade, matemática e biologia teórica. Consiste de uma grelha finita e regular de células, cada uma podendo estar em um número finito de estados, que variam de acordo com regras determinísticas. A grelha pode ser em qualquer número finito de dimensões. O tempo também é discreto e o estado de cada célula no tempo *t* é uma função do estado no tempo *t-1* de um número finito de células na sua vizinhança. Essa vizinhança corresponde a uma determinada seleção de células próximas (podendo eventualmente incluir a própria célula). Todas as células evoluem segundo a mesma regra de atualização, baseada nos valores de suas células vizinhas. Cada vez que as regras são aplicadas à grelha completa, uma nova geração é produzida. Os autômatos celulares foram introduzidos por Von Neumann e Ulam [Wolfram. S, 2002] como modelos para estudar processos de crescimento e auto-reprodução. Qualquer sistema com muitos elementos idênticos que interagem local e deterministicamente podem ser modelados usando autômatos celulares.

Assim, a solução encontrada em [Cao, et al. 1998] para as perturbações das soluções, foi controlar a seleção com base no grid do AC para evitar a perda rápida da diversidade durante a pesquisa genética. Tal estrutura baseia-se na distribuição paralela das populações, onde a idéia de isolar os indivíduos possibilita grande diferenciação genética [S. Wright, 1943]. Em alguns casos [Alba, and Troya, 2002], estes algoritmos utilizando populações descentralizadas conseguem prover uma melhor amostragem do espaço de busca e também melhorar o tempo de execução. Da combinação entre algoritmos genéticos e autômatos celulares surgiu outro tipo de algoritmo evolucionário que tem trazido resultados interessantes, o algoritmo genético celular (AGC), que, por sua vez, pertence a um subconjunto dos algoritmos evolucionários chamados de algoritmos evolucionários celulares (AEC).

Nos AECs o conceito de vizinhança é muito utilizado e um indivíduo deve interagir somente com seus vizinhos mais próximos. Através desta restrição de interação, os AECs conseguem explorar o espaço de busca com maior eficácia, uma vez que a difusão lenta de soluções através da população fornece uma maior *exploration*, enquanto a *exploitation* ocorre dentro de uma vizinhança por meio de operações genéticas [Alba and Dorronsoro, 2008].

O modelo celular simula a evolução natural do ponto de vista do indivíduo. A principal idéia deste modelo é dar a população uma estrutura especial definida como um grafo conectado, no qual cada vértice é um indivíduo que se comunica com os seus vizinhos mais próximos. Ao visualizar a população de algoritmos genéticos desta forma, podemos dizer que em um AG padrão, sua população estaria como um grafo completamente conectado, onde os vértices são os indivíduos e as arestas as suas relações, enquanto num AG celular as interações acontecem apenas com os vizinhos mais próximos como um grafo de estrutura. A figura 1 mostra estes grafos [Alba and Dorronsoro 2008].

Os AGCs são uma subclasse dos algoritmos evolucionários no qual a população é distribuída num grid bidimensional toroidal [D. Simoncini, et al 2007]. Além de possuírem todos os conceitos de um AG comum, os AGCs incrementam ao seu processo evolutivo um autômato celular.

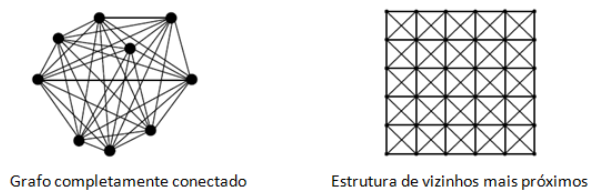

**Figura 1. Grafo completamente conectado e grafo de estrutura**

Cada célula do autômato possui um indivíduo do AG e os operadores genéticos (seleção, reprodução e mutação) são aplicados na vizinhança de um determinado cromossomo escolhido no grid. Como dito anteriormente, a evolução natural aqui se faz voltada para o ponto de vista do indivíduo, ou seja, só há cruzamento entre aqueles indivíduos mais próximos, simulando a influência entre os indivíduos numa sociedade. As interações da população dependem fortemente de sua topologia de vizinhança. A figura 2 mostra os dois tipos de vizinhança mais encontrados na literatura, a vizinhança de Moore e a vizinhança de Neumann.

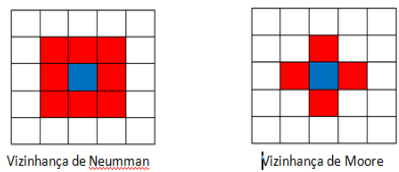

**Figura 2. Tipos de Vizinhanças**

## 4. O método AGCRN-CR

O desenvolvimento do AGCRN-CR utiliza RNAs Evolucionárias (RNAEs). A busca evolucionária é subdividida em camadas, onde cada camada possui uma determinada informação da RNA a ser "evoluída" por um AG específico. No AGCRN-CR, em cada camada, há um AGC que trabalha com a informação pertinente a camada em questão. O uso do AGC visa escapar dos mínimos locais e encontrar redes com melhores configurações em relação a um AG padrão. Essa pesquisa é feita da seguinte forma:

- Na camada mais baixa é feita a busca evolucionária por pesos iniciais da RNA;

- Na camada intermediária ocorre a busca evolucionária por arquiteturas e funções de ativação. Nesta busca também são incluídos o número de camadas escondidas e o número de neurônios por camada;

- Na camada superior a busca evolucionária por regras de aprendizagem (parâmetros dos algoritmos de treinamento) é feita.

A figura 3 mostra uma possível configuração de busca.

Como é possível observar na figura 3, a velocidade de evolução é maior na camada inferior (busca por pesos iniciais). Esta é uma consequência da alta dimensionalidade do espaço de exploração de pesos iniciais devido à falta de conhecimento a priori sobre conjuntos de pesos iniciais. As demais camadas possuem um maior conhecimento a priori e por isso permitem a restrição de espaços de busca mais específicos.

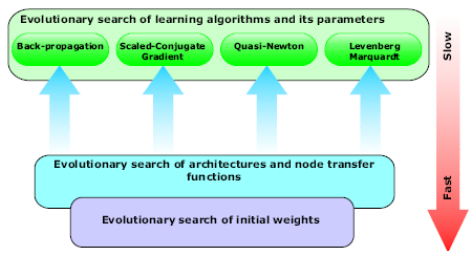 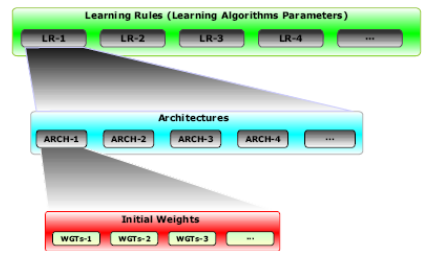

**Figura 3. Busca evolucionária em camadas**     **Figura 4. Composição dos dados na busca evolutiva**

O AGCRN-CR realiza a busca por RNAs totalmente conectadas com aprendizado supervisionado para problemas de classificação que tem uma arquitetura simples (com poucas camadas ocultas e nodos por camada), convergência mais rápida com poucas épocas de treinamento e satisfatória performance de generalização. Por exemplo, no trabalho de Abraham [Abraham 2004] são usadas até 500 épocas de treinamento para cada rede, enquanto o método AGCRN-CR utiliza até 5 épocas. A composição dos dados na busca evolutiva em camadas do AGCRN-CR é mostrada na figura 4.

A figura 4 mostra que cada camada trabalha com uma população diferente de indivíduos que irá codificar um atributo em questão das RNAs. Na camada inferior há a população de pesos iniciais (PPI), na camada intermediária há a população de arquiteturas e funções de ativação (PAF) e na camada superior há a população de regras de aprendizado (PRA). Como já citado anteriormente, cada uma dessas populações trabalha com os valores reais específicos aos seus domínios de problema, ou seja, a PPI possui populações de cromossomos com os dados reais dos pesos das redes neurais, a PAF populações de cromossomos com os dados reais de configuração de arquiteturas e funções de ativação aplicadas e a PRA possui os dados reais dos parâmetros dos algoritmos de aprendizagem aplicados.

Uma vez que os dados reais são utilizados, o processo de codificação e descodificação dos cromossomos é dispensado, mas é preciso adaptar os operadores genéticos para funcionarem com tal tipo de codificação. Estes operadores foram desenvolvidos em [Almeida and Ludermir, 2007] e foram adaptados para trabalhar com os algoritmos genéticos celulares do método AGCRN-CR.

A seleção de $n$ indivíduos é realizada usando a estratégia de torneio, com uma taxa de pressão aleatória $p = 50\%$ e uma taxa de elitismo $e = 10\%$. Esta informação foi encontrada empiricamente. O operador de seleção por torneio é usado tanto para seleção de sobreviventes como para seleção de pais para o cruzamento

Cada PPI está associada a uma arquitetura fixa pré-definida. Dessa forma todos os seus indivíduos possuem as mesmas dimensões. Assim, o cruzamento dessa população ocorre como descrito no algoritmo 1. Os filhos serão formados pelo cruzamento dos pais, que serão divididos ao meio para formar novos indivíduos. Esta idéia visa manter as correlações entre os pesos. Portanto, sejam $a$ e $b$ duas matrizes, as metades $a_d, a_e, b_d$ e $b_e$ são obtidas e irão definir a nova matriz filha, por exemplo, $w_{filho}=[a_d; b_e]$ ou $w_{filho} = [a_e; b_d]$

Uma vez que os indivíduos ainda não possuem uma aptidão, a seleção deste para o processo de mutação é feita aleatoriamente.

**Algoritmo 1**: Operador de cruzamento para indivíduos da PPI
___
**Entrada**: $ppi_n$, $w$ // pais, número de matrizes na rede
**Saída:** $vetFilhos$ // filhos
___
1  **inicio**
2        $numFilhos \leftarrow 1$
3        **enquanto** $numFilhos \leq (n/2)$ **faça**
4            **para** $matriz \leftarrow 1:w$ **faça**
5                $metadeA \leftarrow fatia(ppi_n(numFilhos).(matriz))$
6                $metadeB \leftarrow fatia(ppi_n(numFilhos + 1).(matriz))$
7                $filho(matriz) \leftarrow [metadeA; metadeB]$
8            $vetFilhos \leftarrow filho$
9            $numFilhos = numFilhos + 2$
10 **fim**

A taxa de mutação empregada é de $m = 40\%$, quarenta por cento dos filhos passarão pelo processo de mutação e estes sofrerão mutação em 40% de sua composição. Esta mutação é feita através de matrizes esparsas com 40% de seus valores entre $fx = [-0.5, 0.5]$ e os demais valores definidos como zero. Estas matrizes esparsas serão adicionadas aos filhos gerando a mutação. O fitness dos indivíduos da PPI é avaliado de acordo com o erro médio quadrático normalizado (normalized mean square error – NMSE) do conjunto de treinamento da rede.

Para indivíduos da PAF a aptidão $I_{fit}$ é composta por quatro partes de informação, $I_{acc}$ acurácia de classificação, $I_{nmse}$ erro de treinamento, $I_{comp}$ complexidade da rede e $I_f$ peso da função de transferência utilizada.

$$I_{fit} = \propto * I_{acc} + \beta * I_{nmse} + \gamma * I_{comp} + \delta * I_f \quad (1)$$

$$I_{acc} = 100 * \left(1 - \frac{correct}{total}\right) \quad (2)$$

$$I_{nmse} = \frac{100}{N\,P} * \sum_{j=1}^{P} \sum_{i=1}^{N}(d_i - o_i)^2 \quad (3)$$

$$I_{comp} = \frac{c}{c_{total}} \quad (4)$$

$$I_f = \sum_{h=1}^{n} f_{a_h} \quad (5)$$

Na equação 1, $I_{acc}$ é a porcentagem de erro de classificação; $I_{nmse}$ é o NMSE de treinamento gerado pela rede; $I_{comp}$ é a medida da complexidade, considerando o número de conexões usadas $c$ e o número de conexões possíveis $c_{total}$; $I_f$ calcula o peso das funções de transferências usadas. Para isso, cada função de transferência tem um peso associado empiricamente: Linear de 0.2, Tangente de 0.4 e Logaritmica de 0.7, priorizando funções de transferência simples, uma vez que o objetivo da metodologia é encontrar redes simples com alta performance; $N$ e $P$ o número total de saídas e o número total de padrões de treinamento respectivamente; $d$ e $o$ são as saídas desejas e as saídas obtidas pela rede respectivamente. Como o AGCRN-CR busca redes com até 3 camadas escondidas o valor de $c_{total}$ é baseado no número de nodos de cada camada (entrada, saída e escondidos): $c_{total} = ent + esc + esc + esc + saida$. Neste trabalho o critério de classificação adotado foi o winner-takes-all, no qual o nodo de saída com maior valor vai determinar o padrão de classificação. O calculo do $I_{acc}$ ocorre através do critério winner-takes-all.

As constantes $\alpha, \beta, \gamma$ e $\delta$ possuem valores entre [0,1] e controlam a influência dos fatores sobre o processo global de aptidão. Por exemplo, para favorecer a precisão da classificação quanto ao erro de treinamento e complexidade, estas constantes são definidas da seguinte forma $\alpha = 1$, $\beta = 0.90$, $\gamma = 0.90$ e $\delta = 0.20$. Assim, quando indivíduos aparentemente semelhantes forem encontrados, aquele que tiver o menor erro de treinamento, complexidade estrutural ou complexidade de função de transferência prevalecerá. Estes valores foram encontrados empiricamente.

A busca por arquiteturas é feita para encontrar configurações com até três camadas escondidas e até 12 nodos por camada escondida. A seleção de indivíduos para cruzamento (pais) é feita pelo algoritmo 2, que começa definindo o número de camadas escondidas que os descendentes terão, que será a média arredondada para cima ou para baixo dependendo de certa probabilidade. As dimensões de cada camada escondida e as funções de transferência serão escolhidas aleatoriamente, mas dentro dos limites da união das dimensões e funções de ativação dos pais. Estas etapas do algoritmo 2 se repetem até que o número de cruzamentos desejado seja atingido. O operador de cruzamento da PAF pode produzir indivíduos similares aos seus pais, ocasionado pela simplicidade do operador.

---

**Algoritmo 2**: Operador de cruzamento para indivíduos da PAF

**Entrada**: $paf_n, n, k$ // pais, total de indivíduos, indice
**Saída**: $vetFilhos$ // filhos

1 **inicio**
2     **enquanto** $k \leq (n/2)$ **faça**
3         $totalCamadas \leftarrow paf(k).numCamEsc + paf(k+1).numCamEsc$
4         **se** $rand(1) > probs$ **faça**
5             $filho.numCamEsc \leftarrow ceil(totalCamadas/2)$
6         **senão**
7             $filho.numCamEsc \leftarrow floor(totalCamadas/2)$
8         $dimensoes \leftarrow [paf(numFilhos).dim; paf(numFilhos+1).dim]$
9         $funcoes \leftarrow [paf(numFilhos).func; paf(numFilhos+1).func]$
10        **para** $elementos \leftarrow 1:dimensoes_{tamanho}$ **faça**
11             $[filho.dim, filho.func] \leftarrow sorteia(dimensoes, funcoes)$
12         $vetFilhos \leftarrow filho$
13         $numFilhos \leftarrow numFilhos + 2$
10 **fim**

---

Devido a isso, após o cruzamento é necessário a aplicação do operador de mutação para manter a diversidade da população. O Algoritmo 3 descreve a operação de mutação da PAF. Ele começa com a definição de filhos que sofrerão mutação. Um número aleatório é gerado no intervalo $[0,1]$. Este valor será utilizado como uma probabilidade que indicará se uma arquitetura irá sofrer incremento ou decremento de camadas (com o limite máximo de 3 camadas). O mesmo número será usado para definir se haverá um incremento ou decremento no número de neurônios escondidos por camada (com o máximo de 12 por camada). Se uma arquitetura tem apenas uma camada escondida a chance dessa camada se tornar três é a maior dentre todas as possibilidades. No caso de um valor selecionado aleatoriamente não causa uma mudança significativa nas camadas ocultas, um conjunto de valores entre $arqval = [-2, 5]$ é gerada e adicionada as dimensões atuais da arquitetura, tendo esta no mínimo 1 e no máximo 12 neurônios. Se uma arquitetura tem duas camadas ocultas, a possibilidade de o número ser três é maior com a intenção de reduzir o número de nós escondidos por camada.

A pesquisa evolucionária por regras de aprendizagem acontece com a busca dos parâmetros dos seguintes algoritmos: Back-Propagation(BP), Levenberg-Marquardt (LM), quasi-Newton (QNA) e Gradiente Conjugado Escalonado (SCG). De acordo com o algoritmo de aprendizagem adotado, uma PRA terá parâmetros deste algoritmo para funcionar. Cada PRA tem uma PAF e sua aptidão é medida através da melhor aptidão dentre todas de sua PAF. A seleção do indivíduo a partir da PRA também usa a estratégia de torneio. As taxas de mutação e elitismo são os mesmos apresentados anteriormente para as outras pesquisas evolutivas. Como os indivíduos da PRA tem informações relacionadas a parâmetros do algoritmo de aprendizagem em questão, o seu cruzamento ocorre com a geração de novos valores dentro de um intervalo com base nos valores dos pais. De acordo com uma probabilidade os filhos realizam a mutação adicionando ou subtraindo valores as suas respectivas taxas. O valor da mutação nestas taxas é de quarenta por cento para mais ou para menos, dependendo também de uma probabilidade.

O funcionamento geral do AGCRN-CR é apresentado no algoritmo 4 e a tabela 1 apresenta a lista de parâmetros utilizados.

O método inicia sua busca gerando todos os indivíduos e calculando suas aptidões. Em seguida, os operadores genéticos são utilizados nas camadas em busca de novas soluções. Tais operadores tentam manter a diversidade genética na população utilizando a seleção por torneio, a mutação e o autômato celular, enquanto fazem a evolução das soluções. Como já citado, a manutenção da diversidade é um ponto considerado importante no uso de algoritmos genéticos, visto que tal diversidade evita que a busca caia em pontos de mínimos locais no espaço de pesquisa. Nos algoritmos genéticos esse tipo de comportamento deve ser cuidadosamente evitado, uma vez que a obtenção de super indivíduos pode reduzir as chances de encontrar o ponto de ótimo ou quase ótimo global. A manutenção da diversidade com o uso do autômato celular é feita através da localidade, uma vez que os indivíduos só conseguem cruzar localmente as chances de escapar dos pontos de mínimo local aumentam, diminuindo as chances de convergência genética com uso das restrições de vizinhança impostas pelo AC.

## 5. Resultados

Os experimentos foram realizados visando comparar os resultados do método AGCRN-CR em relação aos mesmos resultados com o NNGA-DCOD. Para isso foram utilizadas bases conhecidas do repositório UCI [D. J. Newman, et al, 1998]. As bases utilizadas foram Cancer com 9 atributos (atb), 699 exemplos (exp) e 2 classes (cla); Vidros com 9 atb, 214 exp e 6 cla; Cavalos com 58 atb, 364 exp e 3 cla; e Diabetes com 8 atb, 768 exp e 2 cla. A cada iteração os dados foram aleatoriamente divididos em metades. Uma das metades foi utilizada como entrada para os algoritmos com 70% de seus dados para treinamento e 30% para o conjunto de validação. A outra metade da base foi utilizada como conjunto de teste. Para determinar se os resultados obtidos possuem realmente significância estatística, foi utilizado um teste F com o auxílio do software BioStat 5.0. O teste F é um teste de hipóteses que é aproximadamente distribuído com cerca de cinco a dez graus de liberdade e rejeita a hipótese nula de que os dois algoritmos têm a mesma taxa de erro com uma significância de 0,05 se $F > 4{,}74$.

**Algoritmo 3**: Operador de mutação para indivíduos da PAF

**Entrada:** $vetFilhos, n, m$ // vetor com os filhos, número de filhos, taxa de mutação
**Saída:** $filhosMut$ // Filhos mutados

```
1  inicio
2      quantidadeFilhos ← ((m/100) * n) faça
3         enquanto quantidadeFilhos > 0 faça
4            probabilidade ← rand(1)
5            filho ← sortear(vetFilhos)
6            se filho.numCamEsc = 1 faça
7               se probabilidade ≥ 0.6 faça
8                   filho ← adicionaCamadas(3)
9               senão se probabilidade ≥ 0.5 então
10                  filho ← adicionaCamadas(2)
11              senão
12                  filho ← filho + sorteia(arqval)
13
14           senão se filho.numCamEsc = 2 então
15              se probabilidade ≥ 0.6 então
16                  filho ← adicionaCamadas(3)
17              senão se probabilidade ≥ 0.5 então
18                  filho ← reduzCamadaPara(1)
19              senão
20                  filho ← filho + sorteia(arqval)
21
22           senão
23              se probabilidade ≥ 0.6 então
24                  filho ← reduzCamadasPara(2)
25              senão se probabilidade ≥ 0.5 então
26                  filho ← reduzCamadasPara(1)
27              senão
28                  filho ← filho + sorteia(arqval)
29           vetFilhos ← vetFilhos − filho
30           filhosMut ← filho
31           quantidadeFilhos ← quantidadeFilhos − 1
32      filhosMut ← FilhosMut + vetFilhos
10 fim
```

Esta metodologia foi utilizada devido ao fato do método usual (teste T de Student) gerar um aumento de erros do tipo 1, onde os resultados são erroneamente considerados diferentes com mais freqüência do que o esperado, dado o nível de confiança utilizado.

O método NNGA-DCOD em [Almeida and Ludermir, 2007] teve melhores resultados que a busca manual e obteve RNAs mais compactas, rápidas e com bom poder de generalização. Neste estudo nós comparamos as soluções obtidas com base na informação de erro de dez conjuntos de testes e treinamentos obtidos pelos métodos AGCRN-CR e NNGA-DCOD na Tabela 2. A média das arquiteturas (arquit.) é baseada na quantidade de neurônios escondidos. O valor do teste F é dado na direita da tabela 2 para todos os casos estudados. De acordo com a tabela 2, ambos os métodos foram capazes de encontrar redes compactas com boa performance, mas o método AGCRN-CR teve resultados estatisticamente melhores que o método NNGA-DCOD em pelo menos metade dos testes para cada base de dados. Nas bases Cancer e Cavalos o AGCRN-CR obteve médias estatisticamente melhores em três dos quatro algoritmos avaliados. Nas bases Diabetes e Vidros o método celular obteve melhores médias em dois dos quatro algoritmos avaliados. Segundo [Alba and Dorronsoro, 2008], os AGCs conseguem explorar o espaço de busca com maior eficácia, uma vez que a difusão lenta das soluções através da população fornece uma maior *exploration*, enquanto a *exploitation* ocorre dentro da vizinhança por meio de operações genéticas.

---

**Algoritmo 4**: Funcionamento geral do AGCRN-CR

**Entrada**: $parametros\ da\ Tabela\ 1$
**Saída**: $Um\ conjunto\ de\ redes\ neurais\ ótimas\ ou\ quase - ótimas$ // novos descendentes

1 **inicio**
2    Gere aleatoriamente as populações de: Pesos iniciais (PPI), Arquiteturas e funções (PAF) e Regras de aprendizagem (PRA)
2    **para cada** $algoritmo\ de\ aprendizagem$ **faça**
3      $Calcule\ a\ aptidão\ para\ todos\ os\ indivíduos$
4      **para cada** $geração\ \in BERA$ **faça**
5        **para cada** $PAF \in PRA\ \textbf{e}\ geração \in BEAFA$ **faça**
6          **para cada** $PPI \in PAF\ \textbf{e}\ geração \in BEP$ **faça**
7            Crie um autômato celular $At$ de acordo com o tamanho da população $P$;
8            Insira em $At$ a população de acordo com o seu fitness;
9            Crie um autômato auxiliar $At_{aux}$ e faça-o igual a $At$
10           Inicie o cruzamento através do AGC;
11           Seja $Q$ um número de filhos previamente determinado
12           **para** $numFilhos \leftarrow 1{:}Q$ **faça**
- Selecione aleatoriamente um indivíduo de $At$ e selecione seus vizinhos por torneio;
- Comece o processo de cruzamento entre os vencedores do torneio para procurar descendentes
- Dependendo de uma probabilidade $m$ aplique a mutação nos descendentes;
- Avalie os descendentes
- Se algum filho gerado for melhor que o pai, substitua o pai pelo filho na mesma posição no grid do autômato $At_{aux}$

13          Faça $At = At_{aux}$
14          Selecione entre todos os indivíduos, inclusive os novos, sobreviventes para próxima rodada
15       Repita os passos da BEP
16      Repita os passos da BEP

13    Selecionar redes quase-ótimas obtidas com cada algoritmo de aprendizagem
10 **fim**

|  | Parâmetros para | Valores |
|---|---|---|
| **AGs** | *Modo de codificação* | Real |
|  | *Taxa de Elitismo* | 10% |
|  | *Taxa de Mutação* | 40% |
|  | *Tipo de seleção* | Torneio |
|  | *Número de indivíduos por população (PRA, PAF e PPI)* | 4, 9, 9 |
|  | *Quantidade de geraçãoes (BERA, BEAFA, BEP)* | 18, 5, 3 |
|  | *Cálculo de aptidão* | Baseado no NMSE do conj. de treinamento |
| **RNAs** | *Tipo da Rede* | MLP feedforward |
|  | *Algoritmos de aprendizagem empregados* | BP, LM, QNA, SCG |
|  | *funções de transferência* | FL (linear), FLS (logarítmica Sigmoid) e FTH (tangente hiperbólica) |
|  | *Número de camada escondidas* | até 3 |
|  | *Número de unidades escondidas* | até 12 |
|  | *Número de épocas de treinamento* | até 5 |
|  | *Intervalo para os pesos iniciais* | [-0.05, 0.05] |
|  | *Função de transferência da última camada* | FL |
| **Algoritmos de aprendzagem** | *BP* |  |
|  | *Taxa de aprendizado* | [0.05, 0.25] |
|  | *Momento* | [0.05, 0.25] |
|  | *LM* |  |
|  | *Taxa de aprendizado* | [0.001, 0.02] |
|  | *QNA* |  |
|  | *Tamanho do passo* | [1.0e-06, 100] |
|  | *Limite do tamanho do passo* | [0.1, 0.6] |
|  | *Fator de escala para determinar a performance* | [0.001, 0.003] |
|  | *Fator de escala para determinar o tamanho do passo* | [0.001, 0.02] |
|  | *SCG* |  |
|  | *Fator de controle para cálculo de aproximação da informação de segunda-ordem* | [0, 0.0001] |
|  | *Fator regulador da falta de definição da matriz hessiana* | [0, 1.0e-06] |

**Tabela 1**

| Problemas / Algoritmos | | AGCRN-CR | | | NNGA-DCOD | | | Teste F |
|---|---|---|---|---|---|---|---|---|
| | | Arquit. | Erros | | Arquit. | Erros | | |
| | | | Treino | Teste | | Treino | Teste | |
| Cancer | BP | 14 | 24.2726 | 24.3001 | 9 | 23.9684 | 23.9810 | ≤ 4.74 |
| | LM | 12.5 | 1.6778 | 2.0790 | 11.6 | 2.3036 | 2.5402 | > 4.74 |
| | SCG | 11.9 | 4.4640 | 4.0842 | 12.5 | 5.2004 | 5.1257 | > 4.74 |
| | QNA | 11.8 | 2.9263 | 2.6734 | 9.1 | 3.6066 | 3.3305 | > 4.74 |
| Diabetes | BP | 7.1 | 23.9847 | 24.1965 | 12.2 | 24.3103 | 24.4491 | > 4.74 |
| | LM | 16.6 | 13.6740 | 17.2658 | 12.3 | 14.1265 | 16.9845 | ≤ 4.74 |
| | SCG | 16.3 | 17.1709 | 20.1215 | 15.7 | 18.6435 | 21.0650 | > 4.74 |
| | QNA | 12 | 18.7571 | 21.0253 | 16.5 | 17.6658 | 20.0718 | ≤ 4.74 |
| Cavalos | BP | 7.5 | 23.0633 | 22.6557 | 9.1 | 24.1395 | 23.9741 | > 4.74 |
| | LM | 17.8 | 8.8731 | 15.8538 | 16.8 | 9.8776 | 15.1914 | ≤ 4.74 |
| | SCG | 17.9 | 16.1976 | 16.4628 | 12.6 | 15.1021 | 15.7953 | > 4.74 |
| | QNA | 4.8 | 16.6637 | 16.1699 | 16.1 | 18.2964 | 17.2823 | > 4.74 |
| Vidros | BP | 14 | 14.1346 | 13.5554 | 9.7 | 14.2789 | 13.8430 | > 4.74 |
| | LM | 17.8 | 10.2865 | 11.0052 | 16.8 | 9.5976 | 11.7114 | > 4.74 |
| | SCG | 17.9 | 12.1718 | 12.0891 | 18.8 | 12.3062 | 11.7663 | ≤ 4.74 |
| | QNA | 14.8 | 12.4402 | 12.6701 | 14.6 | 12.2023 | 11.2883 | ≤ 4.74 |

**Tabela 2**

## 6. Conclusões

Os algoritmos genéticos celulares possuem melhor equilíbrio entre *exploration* e a *exploitation* quando comparados a um AG comum.

Devido a este equilíbrio, os AGCs conseguem escapar mais facilmente dos pontos de mínimo local no espaço de busca, conseguindo atingir, em média, melhores resultados em relação a um AG comum, exatamente como o adotado pelo método NNGA-DCOD. Os resultados experimentais mostraram que na maioria dos casos o método celular obteve médias estatisticamente menores em relação ao NNGA-DCOD. Em trabalhos futuros há a pretensão de aplicar o método celular através de uma implementação paralela, visando acelerar a busca genética, uma vez que a pesquisa obteve resultados semelhantes em termos de tempo.

## 7. Agradecimentos

## 8. Referências

A. Abraham, (2004) "Meta learning evolutionary artificial neural networks". In Neurocomputing, 56, pp. 1-38.

E. Alba, B. Dorronsoro, (2008) "Cellular Genetic Algorithms". In *Operations Research Computer Science Interfaces Series*, 42.

E. Alba, and J. M. Troya, (2002) "Improving flexibility and efficiency by adding parallelism to genetic algorithms". In *Statistics and Computing*, *pp. 91-114*.

L. M. Almeida, T. B. Ludermir, (2007) "Automatically searching near-optimal artificial neural networks". In Proceedings European Symposium on Artificial Neural Networks 2007, pp. 549-554.

Y. J. Cao, H. Q. Wu, (1998) "A Cellular Automata Based Genetic Algorithm and its Application in Machine Design Optimization." In *ICC, pp. 1593 – 1598*.

M. Carvalho, T. B. Ludermir, (2006) "Particle Swarm Optimization of Feed-Foward Neural Networks with Weight Decay". In Proceedings of the Sixth International Conference on Hybrid Intelligent Systems (HIS'06), pp. 5-5.

A. A. Ferreira, T.B. Ludermir (2009) "Genetic Algorithm for Reservoir Computing Optimization". In Proceedings of International Joint Conference on Neural Networks, 2009, pp. 811-815.

E. M. N. Figueiredo, T.B. Ludermir (2014), "Investigating the use of alternative topologies on performance of the PSO-ELM". In Neurocomputing, 127, pp. 4-12.

N. Garcia-Pedrajas, C. Hervas-Martinez, and D. Ortiz-Boyer (2005) "Cooperative coevolution of artificial neural network ensembles for pattern classification." In IEEE Trans. Evolut. Computation: 9(3), pp. 271-302.

G. S. S. Gomes, T. B. Ludermir (2013) "Optimization of the weights and asymmetric activation function family of neural network for time series forecasting". In Expert Systems with Applications, 40, pp. 6438-6446.

R. Linden (2006), "Algoritmos genéticos – Uma importante ferramenta da inteligência computacional", Rio de Janeiro. Brasport, 1ª edição.

F.L. Minku, T.B. Ludermir (2008) "Clustering and co-evolution to construct neural network ensembles: an experimental study". Neural Network, v. 21, pp. 1363-1379.

D. J. Newman, S. Hettich, C.L. Blake, and C.J. Merz, (1998) "UCI repository of machine learning databases" [Online]. Available: http://www.ics.uci.edu/mlearn/MLReposity.html

H. Rajabalipour, H. Haron, M. I. Jambak (2009) "The Improved Genetic Algorithm for Assignment Problems", In ICSPS, pp. 187 - 191.

D. Simoncini, P. Collard, S. Verel, M. Clergue (2007) "On the influence of selection operators on performances in cellular genetic algorithm." In CEC – 2007, *pp.4706-4713*.

S. Wright (1943) "Isolation by distance." In Genetics, 28, pp. 114 – 138

Wolfram. S (2002) "A New Kind of Science", Wolfram Media, ISBN 1-57955-008-8

X. Yao (1999) "Evolving artificial neural networks." In Proceedings of the IEEE, 87(9): pp. 1423-1447.